\documentclass[]{spie}  

\usepackage{mathrsfs}
\usepackage{rotating}
\usepackage{ctable}
\usepackage{bm}
\usepackage{color}
\usepackage{multirow}
\usepackage{floatrow}
\usepackage{blindtext}
\usepackage{cite}
\newfloatcommand{capbtabbox}{table}[][\FBwidth]

\usepackage{amsmath,amsfonts,amssymb}
\usepackage{graphicx}
\usepackage[colorlinks=true, allcolors=blue]{hyperref}

\title{Unsupervised Denoising of Retinal OCT with Diffusion Probabilistic Model}

\author[a]{Dewei Hu}
\author[b]{Yuankai K. Tao}
\author[c]{Ipek Oguz}
\affil[a]{Vanderbilt University, Dept. of Electrical and Computer Engineering}
\affil[b]{Vanderbilt University, Dept. of Biomedical Engineering}
\affil[c]{Vanderbilt University, Dept. of Computer Science, Nashville, TN, USA}

\authorinfo{Send correspondence to Ipek Oguz(ipek.oguz@vanderbilt.edu)}

\begin{document} 
\maketitle

\begin{abstract}
Optical coherence tomography (OCT) is a prevalent non-invasive imaging method which provides high resolution volumetric visualization of retina. However, its inherent defect, the speckle noise, can seriously deteriorate the tissue visibility in OCT. Deep learning based approaches have been widely used for image restoration, but most of these require a noise-free reference image for supervision. In this study, we present a diffusion probabilistic model that is fully unsupervised to learn from noise instead of signal. A diffusion process is defined by adding a sequence of Gaussian noise to self-fused OCT b-scans. Then the reverse process of diffusion, modeled by a Markov chain,  provides an adjustable level of denoising. Our experiment results demonstrate that our method can significantly improve the image quality with a simple working pipeline and a small amount of training data. The implementation is available at \url{https://github.com/DeweiHu/OCT_DDPM}. 
\end{abstract}

\keywords{OCT, denoising, unsupervised, diffusion probabilistic model}

\section{INTRODUCTION}
\label{sec:intro}  
Due to limited spatial-frequency bandwidth, optical coherence tomography (OCT)~\cite{huang1991optical}
has an inherent characteristic of speckle~\cite{schmitt1999speckle}.  
The speckle can severely degrade the image quality by occluding essential anatomical structures such as retinal vessels and thin layers (e.g., the external limiting membrane (ELM)). Hence, despeckling is an urgent pre-processing step for both clinical diagnoses~\cite{chiu2015kernel} and further image analysis~\cite{xiang2018automatic}. The traditional way to reduce speckle noise is to average multiple noisy b-scan acquisitions at the same location~\cite{sander2005enhanced}. This approach requires a large number of repetitions, and the prolonged acquisition time can be problematic for patient comfort; additionally, registration artifacts caused by eye movement can be an issue. Therefore, a denoising algorithm that does not require repeated acquisitions is desirable.  

Many deep learning methods have been investigated for the OCT denoising problem~\cite{ma2018speckle,devalla2019deep,hu2020retinal,mao2019deep,fan2020oct}. Most of these train the model using low-noise reference images, which is not always available in practice. To tackle this lack of ground truth, some unsupervised approaches including Noise2Noise~\cite{lehtinen2018noise2noise} (N2N) and deep image prior~\cite{ulyanov2018deep} (DIP) have been applied to OCT image restoration. Mao \textit{et al.}~\cite{mao2019deep} implement N2N on OCT denoising by using two repeated b-scans as input and target for model training. Fan \textit{et al.}~\cite{fan2020oct} demonstrate that DIP can have the drawback of serious over-smoothing in despeckling results.

In recent research, Ho \textit{et al.} proposed a diffusion probabilistic model~\cite{ho2020denoising} that is used for image synthesis with a performance superior to  generative adversarial networks (GANs)~\cite{dhariwal2021diffusion}. The general idea of the diffusion model is straightforward: Given a natural image, a Markov chain can be formed by adding a small amount of Gaussian noise at each step. Given a sufficient number of steps, a complex data distribution will finally be transformed to a Gaussian. Conversely, given a Gaussian noise image, a meaningful image can be synthesized in a reverse process. In this work, we propose to leverage the diffusion probabilistic model to denoise retinal OCT b-scans. Since the model is learning the speckle pattern instead of the retina appearance, the reference image used for training is not required to be the true noise-free image. In our experiments we apply the self-fusion~\cite{oguz2020self,hu2021life} method to obtain the clean reference image and we train the parameterized Markov chain by variational inference. As the number of reverse steps is adjustable, the algorithm is able to produce different levels of denoising results. This is an advantage since different tasks may require different levels of fine detail retention in the images.

\section{METHOD}
\label{method}
\begin{figure}[t]
    \centering
    \includegraphics[width=.95\linewidth]{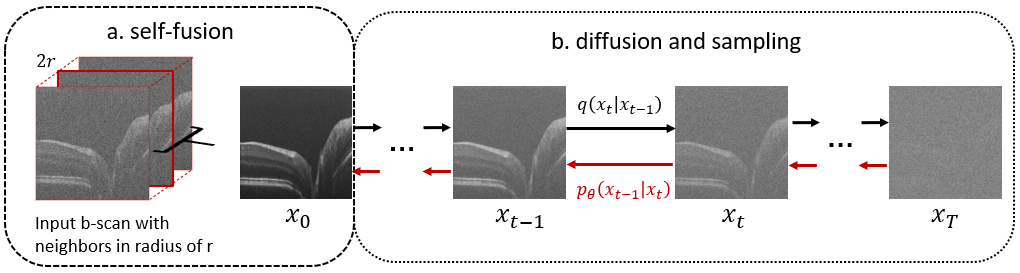}
    \caption{General workflow in this study. In (a) the target b-scan is highlighted in red. The surrounding b-scans within a radius $r$ are used as atlases for reconstructing the target b-scan with higher SNR. In (b) the black arrows indicate the diffusion direction while the red represent sampling.}
    \label{fig:workflow}
\end{figure}

We leverage our self-fusion method~\cite{oguz2020self} as a pre-processing step in the training stage (Fig.~\ref{fig:workflow}a). Self-fusion regards b-scans in a small vicinity of a given target b-scan as `atlases' for that b-scan, because of their structural similarity. After registering the neighbors to the target b-scan, a pixel-wise weighted average of these `atlases' will result in an image with high signal-to-noise ratio (SNR). This approach is easy to implement and robust for retinal layer enhancement, but finer features like vessels and texture can be over-smoothed. Nevertheless, since the diffusion probabilistic model aims to learn the speckle pattern instead of the signal, the self-fusion output $x_0$ can still be used as the clean image for our training purposes. Fig.~\ref{fig:workflow}b shows a Markov chain in forward (diffuse) and reverse (denoise) directions: 
\begin{equation}
    \label{joint}
    q(x_{1:T}|x_0)=\prod_{t=1}^Tq(x_t|x_{t-1}), \qquad p_{\theta}(x_{0:T})=p(x_T)\prod_{t=1}^Tp_{\theta}(x_{t-1}|x_t)
\end{equation}
where $q(x_0)$ represents the data distribution while $p(x_T)=\mathcal{N}(x_t;\bm{0},\bm{I})$. $\theta$ is the parameters of the model. The diffusion and sampling describe a transition between these two distributions with $T$ discretized steps. Our goal is to train a deep model $p_{\theta}$ to restore a noisy image $x$ with an adjustable parameter $t$. Intuitively, an image with stronger speckle require a larger $t$ value that indicates more denoising steps.

The image sequence $x_0,x_1,\hdots,x_T$ is created by gradually adding small Gaussian noise with a variance schedule $\{\beta_1,\hdots,\beta_T\}$, where $\beta_t\in(0,1)$, $\forall t\in(1,T)$. 
\begin{equation}
    \label{df1}
    q(x_t|x_{t-1}):=\mathcal{N}(x_t;\sqrt{\alpha_t}x_{t-1},\beta_t\bm{I}) \quad \textrm{where} \quad \alpha_t = 1-\beta_t
\end{equation}
Eq.~\ref{df1} approximates the posterior distribution in the forward process assuming that there is a small mean shift after one step of diffusion. Denote $\Bar{\alpha}_t=\prod_{s=1}^t\alpha_s$, then $x_t$ is acquired by adding $t$ different Gaussian random variables to $x_0$. 
\begin{equation}
    \label{df2}
    q(x_t|x_0)=\mathcal{N}(x_t;\sqrt{\Bar{\alpha}_t}x_0,(1-\Bar{\alpha}_t)\bm{I})
\end{equation}
As the sum of $t$ Gaussians is a Gaussian with variance $\sum_{t=1}^T\beta_t\approx 1-\Bar{\alpha}_t$. The high order terms with regard to $\beta_t$ in $1-\Bar{\alpha}_t$ are negligible because $\beta_t$ is a small value in range $(0,1)$. In practice, Eq.~\ref{df2} enables sampling of $x_t$ with reparameterization:
\begin{equation}
    \label{df21}
    x_t(x_0,\epsilon) = \sqrt{\Bar{\alpha}_t}x_0 + \sqrt{1-\Bar{\alpha}_t}\epsilon, \quad \epsilon \in \mathcal{N}(\bm{0},\bm{I})
\end{equation}
Similar to Eq.~\ref{df1}, the reverse step is also modeled as a Gaussian since the noise added in each step is small:
\begin{equation}
    \label{dn1}
    p_{\theta}(x_{t-1}|x_t)=\mathcal{N}(x_{t-1};\bm{\mu}_{\theta}(x_t,t),\bm{\Sigma}_{\theta}(x_t,t))
\end{equation}
In this work, we set the variance to be a fixed parameter $\bm{\Sigma}_{\theta}(x_t,t)=\beta_t$ and only learn to predict the mean. 

To perfectly recover the image in the reverse process, the ideal solution is to minimize the distance between $q(x_{t-1}|x_t)$ and $p(x_{t-1}|x_t)$. However, according to Ho \textit{et al.}~\cite{ho2020denoising}, the direct KL-divergence between these two distributions, $D_{KL}(q(x_{t-1}|x_t)||p_{\theta}(x_{t-1}|x_t))$, is not tractable. Alternatively, they introduce a tractable constraint $D_{KL}(q(x_{t-1}|x_t,x_0)||p_{\theta}(x_{t-1}|x_t))$. By leveraging the property of Markov chain, the following equation should hold:
\begin{equation}
    q(x_{t-1}|x_t,x_0) =\frac{q(x_t|x_{t-1},x_0)q(x_{t-1}|x_0)}{q(x_t|x_0)} = q(x_t|x_{t-1})q(x_{t-1}|x_0)
\end{equation}
From Eq.~\ref{df1} and Eq.~\ref{df2}, we know that both terms in the product are Gaussian; then $q(x_{t-1}|x_t,x_0)$ should also have the form  $\mathcal{N}(x_{t-1};\bm{\Tilde{\mu}}_t(x_t,x_0),\Tilde{\beta_t}\bm{I})$:
\begin{equation}
    \label{df3}
    q(x_{t-1}|x_t,x_0) =\mathcal{N}\left(x_{t-1}; \underbrace{\frac{\sqrt{\Bar{\alpha}_{t-1}}\beta_t}{1-\Bar{\alpha}_t}x_0+\frac{\sqrt{\alpha_t}(1-\Bar{\alpha}_{t-1})}{1-\Bar{\alpha}_t}x_{t}(x_0,\epsilon)}_{\bm{\Tilde{\mu}}_t(x_t,x_0)},\underbrace{\frac{1-\Bar{\alpha}_{t-1}}{1-\Bar{\alpha}_t}\beta_t}_{\Tilde{\beta_t}}\bm{I}\right), \quad \epsilon \in \mathcal{N}(\bm{0},\bm{I})
\end{equation}
We use the negative log likelihood $-\log p_{\theta}(x_0)$ as the objective function. This can be optimized by minimizing its variational upper bound $\mathcal{L}$ given by the Jensen's inequality; the detailed derivation of this is included in Appendix \ref{append1}:
\begin{equation}
    \mathcal{L}=\underbrace{D_{KL}(q(x_T|x_0)\ \| \ p(x_T))}_{\mathcal{L}_T}+\sum_{t=2}^T\underbrace{ D_{KL}(q(x_{t-1}|x_t,x_0)\ \| \ p_{\theta}(x_{t-1}|x_t))}_{\mathcal{L}_{t-1}}+\underbrace{\mathcal{H}(p_{\theta}(x_0|x_1))}_{\mathcal{L}_0}
\end{equation}
Because the variance schedule is fixed in this implementation, $\mathcal{L}_T$ turns out to be a constant, so we only need to consider $\mathcal{L}_{t-1}$ and $\mathcal{L}_0$ as loss function. Given Eq.~\ref{df3} and Eq.~\ref{dn1}, $\mathcal{L}_{t-1}$ is the KL divergence of two Gaussian distributions and can be reduced to Eq.~\ref{loss}. The derivation details can be found in Appendix \ref{append2}.
\begin{equation}
    \label{loss}
    \mathcal{L}_{t-1}=\frac{1}{2\beta_t}\left\|\bm{\Tilde{\mu}}_t(x_t,x_0)-\bm{\mu}_{\theta}(x_t(x_0,\epsilon),t)\right\|^2
\end{equation}
Obviously, to minimize $\mathcal{L}_{t-1}$ we can set the mean prediction equal to $\bm{\Tilde{\mu}}_t(x_t,x_0)$ which is derived from Eq.~\ref{df3} and Eq.~\ref{df21}:
\begin{equation}
    \label{express}
    \bm{\mu}_{\theta}(x_t,t) = \frac{1}{\sqrt{\alpha_t}}\left(x_t-\frac{\beta_t}{\sqrt{1-\Bar{\alpha_t}}}\epsilon_{\theta}(x_t,t)\right)
\end{equation}
Combining Eq.~\ref{loss} and Eq.~\ref{express} it is easy to see that the model is learning to predict the sample $\epsilon$ drawn from a normal distribution with mean square error (MSE). The expression of $\bm{\mu}_{\theta}(x_t,t)$ demonstrates that the image restoration is actually done by adding a Gaussian noise.

\section{EXPERIMENTS}
\subsection{Dataset}
We train our model on 6 optic nerve head (ONH) volumes of the human retina, and test it on 6 fovea volumes. Each volume contains 500 b-scans of $512\times 500$ pixels. To test the performance of the model at different speckle levels, the data is acquired with three different levels of signal-to-noise ratio (92dB, 96dB, 101dB). As mentioned in Sec.~\ref{sec:intro}, the ground truth used for evaluation is often obtained by averaging repeated b-scan frames at the same spatial location. In our dataset, we have 5 repeated acquisitions for each b-scan which are averaged together to form the ground truth.

\subsection{Implementation details}
For self-fusion, we take adjacent b-scans within radius of 3 as candidates to compute the weighted mean. We use the \textit{greedy}\footnote{https://github.com/pyushkevich/greedy} software for diffeomorphic image registration~\cite{yushkevich2016fast}. Images are padded to $512\times 512$ pixels, and the intensity is normalized to $[-1,1]$. 

The variance schedule is set to linearly increase from $10^{-4}$ to $6\times 10^{-3}$ in $T=100$ steps. The architecture of our model is a simplified version of that in Ho \textit{et al.}~\cite{ho2020denoising}. It is trained on an NVIDIA RTX 2080TI 11GB GPU for 500 epochs, with a batch size of 2 using Adam optimizer. The starting learning rate is $10^{-4}$ and decay by half every 5 epochs.

\fboxrule=1.2pt
\fboxsep=0.3mm

\begin{figure}[t]
\centering
\begin{tabular}{cccc}
 
  & \large \textbf{SNR=92dB} & \large \textbf{SNR=96dB} & \large \textbf{SNR=101dB} \\
 \rotatebox{90}{\hspace{1cm}\large \textbf{t=0}} &
 {\includegraphics[width=0.28\linewidth]{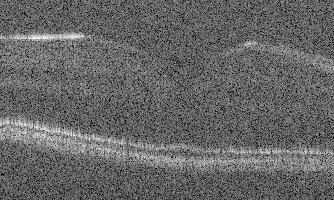}} &
 {\includegraphics[width=0.28\linewidth]{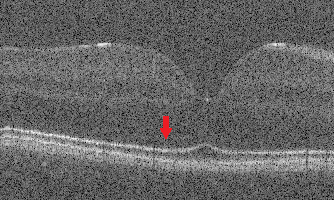}} &
 {\includegraphics[width=0.28\linewidth]{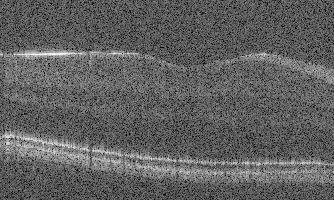}}  \\
 
 \rotatebox{90}{\hspace{1cm}\large \textbf{t=20}}&
 {\includegraphics[width=0.28\linewidth]{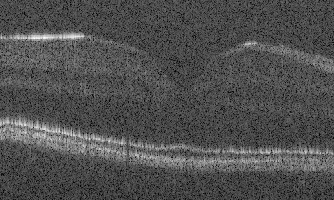}} &
 {\includegraphics[width=0.28\linewidth]{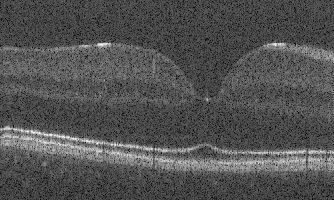}} &
 {\includegraphics[width=0.28\linewidth]{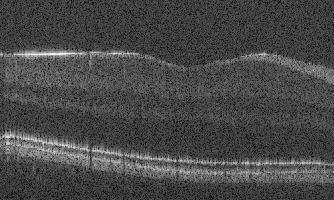}}  \\
 
 \rotatebox{90}{\hspace{1cm}\large \textbf{t=41}} &
 {\includegraphics[width=0.28\linewidth]{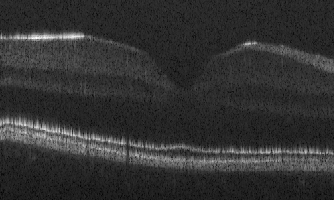}} &
 {\includegraphics[width=0.28\linewidth]{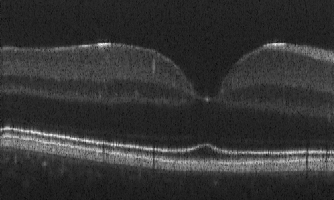}} &
 \fcolorbox{red}{white}{\includegraphics[width=0.28\linewidth]{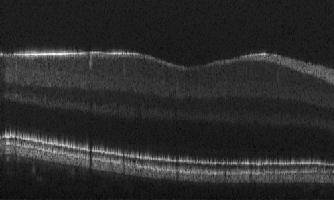}}  \\
 
 \rotatebox{90}{\hspace{1cm}\large \textbf{t=46}} &
 {\includegraphics[width=0.28\linewidth]{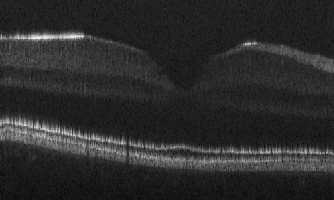}} &
 \fcolorbox{red}{white}{\includegraphics[width=0.28\linewidth]{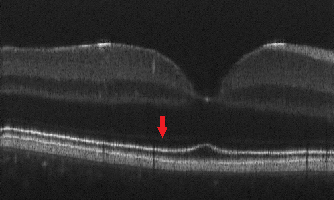}} &
 {\includegraphics[width=0.28\linewidth]{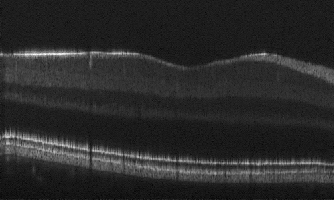}}  \\
 
 \rotatebox{90}{\hspace{1cm}\large \textbf{t=51}} &
 \fcolorbox{red}{white}{\includegraphics[width=0.28\linewidth]{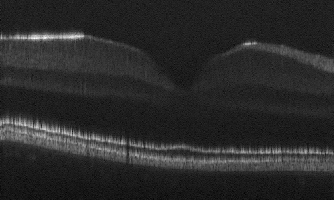}} &
 {\includegraphics[width=0.28\linewidth]{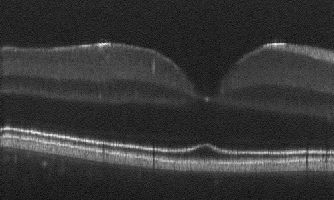}} &
 {\includegraphics[width=0.28\linewidth]{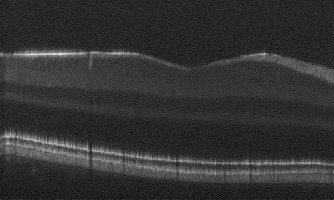}}  \\
 
 \rotatebox{90}{\hspace{1cm}\large \textbf{t=70}} &
 {\includegraphics[width=0.28\linewidth]{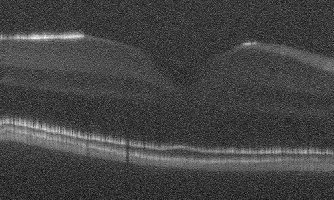}} &
 {\includegraphics[width=0.28\linewidth]{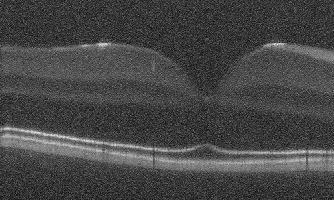}} &
 {\includegraphics[width=0.28\linewidth]{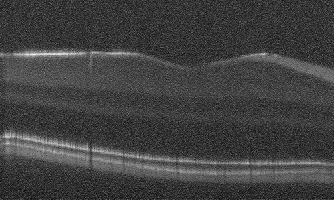}}  \\
 
\end{tabular}

\caption{Fovea denoising results for different input SNR levels and for different t values. The best result (determined visually) for each SNR level is highlighted with {\bf a red box}. The {\bf red arrows} point to the ELM, which becomes visible with our denoising model.  (Excess background trimmed.)}
\label{fig:t-spectrum}
\end{figure}

\section{RESULTS}
\subsection{Qualitative results}
In Fig.~\ref{fig:t-spectrum} we show the denoising results of our model for a range of $t$ values. $t=0$ is the original input image for each SNR level. Increasing $t$ values indicate more denoising steps. The best result (determined visually) for each noise level, as highlighted by the red box, coincide with the intuition that the noisier images benefit from a larger $t$. For the third column, where the input noise level is relatively low, we can see that as $t$ increases from 41 to 51, retinal layers gradually become over-smoothed and fine texture features fade away. As discussed in Sec.~\ref{method}, Eq.~\ref{express} explains that the denoising process is done by adding Gaussian noise to compensate for the speckle pattern. When the $t$ is too large (e.g.~$t=70$ in Fig.~\ref{fig:t-spectrum}), the added noise becomes excessive and produces poor results. 

We further note that our proposed method performs well for vessel and layer preservation. For example, in the second column, our result reveals the very thin external limiting membrane (ELM) (marked by red arrows) which is hardly visible in the noisy input.

\subsection{Comparison to baseline denoising model}
\begin{figure}[p]
\centering
\begin{tabular}{cccc}
 
  & \large \textbf{SNR=92dB} & \large \textbf{SNR=96dB} & \large \textbf{SNR=101dB} \\
 \rotatebox{90}{\hspace{0.7cm} \textbf{5-mean}} &
 {\includegraphics[width=0.28\linewidth]{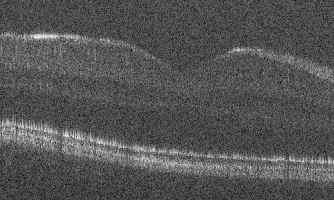}} &
 {\includegraphics[width=0.28\linewidth]{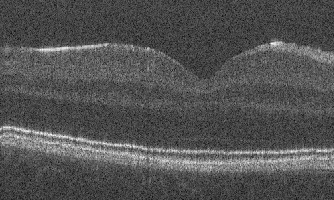}} &
 {\includegraphics[width=0.28\linewidth]{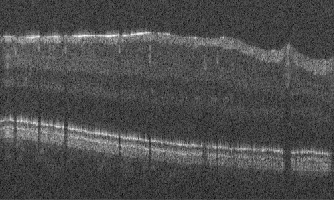}}  \\
 
 \rotatebox{90}{\hspace{0.7cm} \textbf{PMFN}}&
 {\includegraphics[width=0.28\linewidth]{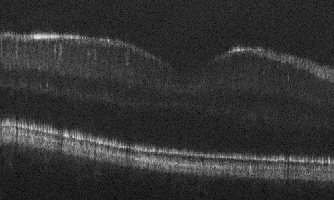}} &
 {\includegraphics[width=0.28\linewidth]{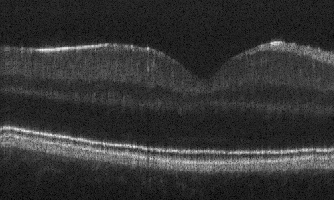}} &
 {\includegraphics[width=0.28\linewidth]{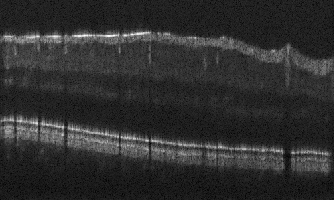}}  \\
 
 \rotatebox{90}{\hspace{0.5cm} \textbf{Proposed}} &
 {\includegraphics[width=0.28\linewidth]{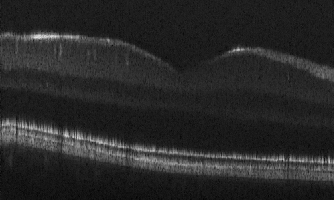}} &
 {\includegraphics[width=0.28\linewidth]{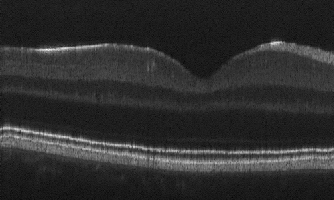}} &
 {\includegraphics[width=0.28\linewidth]{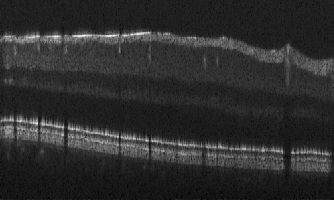}}  \\
 
\end{tabular}

\caption{Result comparison with baseline model. 5-mean refers to the average image of 5 repeated b-scans.}
\label{fig:result}
\end{figure}

Hu \textit{et al.}~\cite{hu2020retinal} present a pseudo-modality fusion network (PMFN) that improves the feature preservation compared to a former method developed by Devalla \textit{et al.}~\cite{devalla2019deep}. We use the PMFN as the baseline in this study. In Fig.~\ref{fig:result}, we observe that the retinal layers are more homogeneous in our proposed approach than in PMFN for all input SNR levels. Downstream analysis tasks such as layer segmentation would likely benefit from this improvement. We also note that small features like vessels are not sacrificed, even though other regions of the layers become denoised. To quantitatively confirm these observations, we use the average of 5 repeated frames (5-mean) as the reference ground truth image and we report several metrics in Table \ref{tab:eval}. Comparing with PMFN, the proposed method improve the denoising performance in terms of SNR, CNR and ENL. The results are significantly different in a paired, two-tailed t-test with a significance threshold of 0.05. 

\begin{table}[h]
\centering
\scalebox{1.0}{
  \begin{tabular}{l|l|l|l|l}
  \specialrule{.1em}{.05em}{.05em}
     & SNR & PSNR & CNR & ENL \\
    \specialrule{.1em}{.05em}{.05em}
    PMFN & $29.18\pm 2.03$ & $81.51\pm 0.69$ & $1.89\pm 0.57$ & $10.91\pm 2.80$ \\
    proposed & $\bm{40.94\pm 1.78}$ & $74.67\pm 0.58$ & $\bm{2.12\pm0.71}$ & $\bm{54.66\pm 15.84}$\\
    \specialrule{.1em}{.05em}{.05em}
  \end{tabular}}
  \caption{Quantitative evaluation. SNR: signal-to-noise ratio, PSNR: peak signal-to-noise ratio, CNR: contrast-to-noise ratio, ENL: equivalent number of looks. Improvements over the baseline are highlighted in boldface.}
  \label{tab:eval}
\end{table}

\section{CONCLUSION}
We propose an OCT restoration method leveraging a diffusion probabilistic model. The model is unsupervised and requires a small amount of data to train. Moreover, the parameter $t$ provides control over the output smoothness level. Our results show that our model can efficiently suppress the speckle; furthermore, the features including layers and vessels are not only preserved but present higher visibility. A limitation of this approach is its Gaussian assumption on the speckle pattern. The generalization to other noise distribution types will be a potential direction of future work for better speckle modeling.

\section{ACKNOWLEDGEMENTS}
This work is supported by NIH R01EY031769, NIH R01EY030490 and the Vanderbilt University Discovery Grant Program.

\clearpage

\bibliography{report} 
\bibliographystyle{spiebib} 

\appendix
\section{variational upper bound}
\label{append1}
According to the Jensen's inequality, for a concave function $f$, we should have:
\begin{equation}
    f(\mathbb{E}[x])\geq \mathbb{E}[f(x)]
\end{equation}
Hence the log likelihood $\log p_{\theta}(x_0)$ has a evidence lower bound (ELBO), as logarithm is a concave function:
\begin{align*}
    \log p_{\theta}(x_0)&=\log \int_{x_{1:T}}p_{\theta}(x_0|x_{1:T})p(x_{1:T}) \\
    & = \log \int_{x_{1:T}}p_{\theta}(x_{0:T}) \frac{q(x_{1:T}|x_0)}{q(x_{1:T}|x_0)}\\
    & = \log \int_{x_{1:T}}\left[\frac{p_{\theta}(x_{0:T})}{q(x_{1:T}|x_0)}\right]q(x_{1:T}|x_0) \\
    & = \log \mathbb{E}_{x_{1:T}\sim q(x_{1:T}|x_0)}\left[\frac{p_{\theta}(x_{0:T})}{q(x_{1:T}|x_0)}\right] \\
    &\geq \mathbb{E}_{x_{1:T}\sim q(x_{1:T}|x_0)}\left[\log \frac{p_{\theta}(x_{0:T})}{q(x_{1:T}|x_0)}\right]
\end{align*}
The negative log likelihood will then have an upper bound:
\begin{align*}
    -\log p_{\theta}(x_0) &\leq \mathbb{E}_{x_{1:T}\sim q(x_{1:T}|x_0)}\left[-\log \frac{p_{\theta}(x_{0:T})}{q(x_{1:T}|x_0)}\right]\\
    & = \mathbb{E}_{x_{1:T}\sim q(x_{1:T}|x_0)}\left[-\log \frac{p(x_T)\prod_{t=1}^T p_{\theta}(x_{t-1}|x_t)}{\prod_{t=1}^T q(x_t|x_{t-1})}\right]\\
    & = \mathbb{E}_{x_{1:T}\sim q(x_{1:T}|x_0)}\left[-\log p(x_T)\prod_{t=1}^T\frac{p_{\theta}(x_{t-1}|x_t)}{q(x_t|x_{t-1})}\right]\\
    & = \mathbb{E}_{x_{1:T}\sim q(x_{1:T}|x_0)}\left[-\log p(x_T)-\sum_{t=1}^T\log\frac{p_{\theta}(x_{t-1}|x_t)}{q(x_t|x_{t-1})}\right]
\end{align*}
Then minimizing the negative log likelihood is equivalent to minimize the upper bound $\mathcal{L}$.
\begin{align*}
    \mathcal{L}&=\mathbb{E}_q\left[-\log p(x_T)-\sum_{t=1}^T\log\frac{p_{\theta}(x_{t-1}|x_t)}{q(x_t|x_{t-1})}\right]\\
    &=\mathbb{E}_q\left[-\log p(x_T)-\sum_{t=2}^T\log\frac{p_{\theta}(x_{t-1}|x_t)}{q(x_t|x_{t-1})}-\log\frac{p_{\theta}(x_{0}|x_1)}{q(x_1|x_{0})}\right]
\end{align*}
Intuitively $q(x_t|x_{t-1})$ can be approximated by $q(x_t|x_{t-1},x_0)$.
\begin{equation*}
    q(x_t|x_{t-1})\approx q(x_t|x_{t-1},x_0)=\frac{q(x_{t-1}|x_t,x_0)q(x_t|x_0)}{q(x_{t-1}|x_0)}
\end{equation*}
hence the bound $\mathcal{L}$ can be expressed as:
\begin{align*}
    \mathcal{L}&=\mathbb{E}_q\left[-\log p(x_T)-\sum_{t=2}^T\log\frac{p_{\theta}(x_{t-1}|x_t)}{q(x_{t-1}|x_{t},x_0)}\frac{q(x_{t-1}|x_0)}{q(x_t|x_0)}-\log\frac{p_{\theta}(x_{0}|x_1)}{q(x_1|x_{0})}\right]\\
    &=\mathbb{E}_q\left[-\log p(x_T)-\sum_{t=2}^T\log\frac{p_{\theta}(x_{t-1}|x_t)}{q(x_{t-1}|x_{t},x_0)}-\log\frac{q(x_{1}|x_0)}{q(x_T|x_0)}-\log\frac{p_{\theta}(x_{0}|x_1)}{q(x_1|x_{0})}\right]\\
    &= \mathbb{E}_q\left[\log\frac{q(x_T|x_0)}{p(x_T)}+\sum_{t=2}^T\log\frac{q(x_{t-1}|x_t,x_0)}{p_{\theta}(x_{t-1}|x_t)}-\log p_{\theta}(x_0|x_1)\right]\\
    &= \underbrace{D_{KL}(q(x_T|x_0)\ \| \ p(x_T))}_{\mathcal{L}_T}+\sum_{t=2}^T\underbrace{ D_{KL}(q(x_{t-1}|x_t,x_0)\ \| \ p_{\theta}(x_{t-1}|x_t))}_{\mathcal{L}_{t-1}}+\underbrace{\mathcal{H}(p_{\theta}(x_0|x_1))}_{\mathcal{L}_0}
\end{align*}

\section{KL divergence of Gaussian variables}
\label{append2}
Let $p(x)$ and $q(x)$ be two Gaussian distributions:
\begin{align*}
    p(x) \sim \mathcal{N}(\mu_1,\sigma_1^2),\  q(x) \sim \mathcal{N}(\mu_2,\sigma_2^2)
\end{align*}
 The KL divergence between $p(x)$ and $q(x)$ can be derived as follows:
\begin{align*}
    D_{KL}(p(x)\ \| \ q(x)) &= \mathbb{E}_p\left[\log\frac{p(x)}{q(x)}\right] = \int p(x)\log p(x)dx-\int p(x)\log q(x)dx \\
    & = -\frac{1}{2}(1+\log 2\pi\sigma_1^2)+\frac{1}{2}\log 2\pi\sigma_2^2+\frac{1}{2\sigma_2^2}\left[\sigma_1^2+(\mu_1-\mu_2)^2\right]\\
    &= \frac{1}{2\sigma_2^2}\left[\sigma_1^2+(\mu_1-\mu_2)^2\right] + \log \frac{\sigma_2}{\sigma_1}-\frac{1}{2}
\end{align*}
Apply this result on $\mathcal{L}_{t-1}$, if the ratio of variances is either cancelled or regarded as constant
\begin{equation}
    \mathcal{L}_{t-1}=\frac{1}{2\sigma_t^2}\|\bm{\Tilde{\mu_t}}(x_t,x_0)-\bm{\mu}_{\theta}(x_t,t)\|^2+C
\end{equation}
From the reparameterization (Eq.~\ref{df21}) it is easy to get 
$$x_0=\frac{1}{\sqrt{\Bar{\alpha_t}}}\left(x_t(x_0,\epsilon)-\sqrt{1-\Bar{\alpha_t}}\epsilon\right)$$
Then plug this in $\bm{\Tilde{\mu_t}}$ to get:
\begin{align*}
    \bm{\Tilde{\mu_t}}(x_t,x_0) &= \bm{\Tilde{\mu_t}}\left(x_t(x_0,\epsilon),\frac{1}{\sqrt{\Bar{\alpha_t}}}(x_t(x_0,\epsilon)-\sqrt{1-\Bar{\alpha_t}}\epsilon)\right) \\
    &= \frac{\sqrt{\Bar{\alpha}_{t-1}}\beta_t}{1-\Bar{\alpha}_t} \cdot \frac{1}{\sqrt{\Bar{\alpha}}_t}(x_t(x_0,\epsilon)-\sqrt{1-\Bar{\alpha_t}}\epsilon)+\frac{\sqrt{\alpha_t}(1-\Bar{\alpha}_{t-1})}{1-\Bar{\alpha}_{t}}x_t(x_0,\epsilon) \\
    &= \left[\frac{\beta_t}{(1-\Bar{\alpha}_t)\sqrt{\alpha_t}}+\frac{\sqrt{\alpha_t}(1-\Bar{\alpha}_{t-1})}{1-\Bar{\alpha}_{t}}\right]x_t(x_0,\epsilon)-\frac{\beta_t}{(1-\Bar{\alpha}_t)\sqrt{\alpha_t}}\sqrt{1-\Bar{\alpha_t}}\epsilon \\
    &= \frac{1}{\sqrt{\alpha_t}}\left(\frac{\beta_t+\alpha_t(1-\Bar{\alpha}_{t-1})}{1-\Bar{\alpha}_{t}}x_t(x_0,\epsilon)-\frac{\beta_t}{\sqrt{1-\Bar{\alpha_t}}}\epsilon\right) \\
    &= \frac{1}{\sqrt{\alpha_t}}\left(x_t(x_0,\epsilon)-\frac{\beta_t}{\sqrt{1-\Bar{\alpha_t}}}\epsilon\right)
\end{align*}
Then apply this result in the expression of $\mathcal{L}_{t-1}$, we get:
\begin{equation}
    \mathcal{L}_{t-1}-C=\frac{1}{2\sigma_t^2}\left\|\frac{1}{\sqrt{\alpha_t}}\left(x_t(x_0,\epsilon)-\frac{\beta_t}{\sqrt{1-\Bar{\alpha_t}}}\epsilon\right)-\bm{\mu}_{\theta}(x_t(x_0,\epsilon),t)\right\|^2
\end{equation}
To minimize $\mathcal{L}_{t-1}$, given fixed $x_t$, the following should hold:
\begin{align}
    \bm{\mu}_{\theta}(x_t,t) &= \frac{1}{\sqrt{\alpha_t}}\left(x_t-\frac{\beta_t}{\sqrt{1-\Bar{\alpha_t}}}\epsilon_{\theta}(x_t,t)\right)
\end{align}

\end{document}